# Arabic Named Entity Recognition using Word Representations

*Ismail El bazi[1] and Nabil Laachfoubi*
Univ Hassan 1, IR2M Laboratory, 26000 Settat, Morocco
ismailelbazi@gmail.com

*Abstract*-Recent work has shown the effectiveness of the word representations features in significantly improving supervised NER for the English language. In this study we investigate whether word representations can also boost supervised NER in Arabic. We use word representations as additional features in a Conditional Random Field (CRF) model and we systematically compare three popular neural word embedding algorithms (SKIP-gram, CBOW and GloVe) and six different approaches for integrating word representations into NER system. Experimental results show that Brown Clustering achieves the best performance among the six approaches. Concerning the word embedding features, the clustering embedding features outperform other embedding features and the distributional prototypes produce the second best result. Moreover, the combination of Brown clusters and word embedding features provides additional improvement of nearly 10% in F1-score over the baseline.

*Keywords:* Named Entity Recognition, Word Representations, Word Embeddings, Arabic

## I. INTRODUCTION

In order to achieve good performance, supervised NER models require huge amounts of manually annotated data. The annotation process is fastidious and needs much time and resources. An effective way of handling this data sparsity is to take advantage of massive unlabeled data, freely available, to learn word representations and use it as features to boost supervised NER systems.

The first type of word representations used in NER was Brown clustering. Liang[1] included the Brown cluster features in semi-supervised English NER and achieved substantial improvements. More recently, the focus has switched to a new type of word representations named "word embeddings". The effectiveness of word embedding features in NER has been widely demonstrated for English language. The works of Turian et al. [2], [3], Passos et al. [4] and Guo et al. [5] show that plugging word embeddings into linear models is the key to improve NER and create state-of-the-art systems.

Inspired by the success of English language, we investigate in this paper whether these techniques can be successfully applied to NER in Arabic.

In section 2, first we survey prior work on Arabic NER and present the various methods of learning word representations. Section 3 describes the approaches of integrating word representations into NER. Section 4 outlines the experimental settings and results. Finally, we draw our final conclusions in section 5.

## II. RELATED WORK

### A. Arabic NER

Recently, numerous research studies have been published on the supervised Arabic NER. Most of these NER systems used classical features such as lexical (prefixes, suffixes, character n-grams, word length, and punctuation) [6–8], contextual (word n-grams and rule-based features) [8], [9], morphological (Part of Speech, Gender, Person, Aspect, Number, Base Phrase Chunk, etc.) [10], [11] and list lookup features (gazetteers, lexical Triggers and Nationality information) [9], [11].

---
[1] Corresponding author

The first Arabic NER system that incorporates word representations is the one proposed by Zirikly and Diab [12]. This Dialectal Arabic NER system includes Brown Clustering as feature in addition to classical features and shows improvement over state-of-the-art features performance.

Zirikly and Diab [13] continue the work on word representations for Arabic NER and study the impact of word representations on Arabic NER performance for Social Media data. Their system uses Brown Clustering and Word2vec Clustering Embeddings with Lexical, Contextual, Morphological Features and Gazetteers and demonstrates superior results in comparison to other NER systems using large gazetteers.

*B. Word Representations*

A word representation can be defined as a vector paired with a word, where each dimension's value matches a feature and could potentially capture useful semantic and syntactic properties [2].

Recently, word Representations have been quite successful at substantially improving performance on various NLP tasks [3], [14–16].

There are two main approaches to inducing unsupervised word representations over unlabeled data. One approach is to use clustering algorithms to induce clusters from unlabeled corpora and using them as features in supervised NLP systems. The clustering can be hierarchical like Brown Clustering [1], [17] or non-hierarchical like k-means [18] and Clarke Clustering [19].

Another approach is to learn dense low-dimensional real-valued vectors also known as a "word embeddings" [20] using unsupervised approaches. The two predominant model's families for inducing word embeddings are:

*1) Count-based methods*, such as Latent Semantic Analysis (LSA) [21], Hyperspace Analogue to Language (HAL) [22], Hellinger PCA (HPCA) [23] and Global Vectors (GloVe) [24]. These methods use co-occurrence matrix and matrix factorization techniques to generate word vectors.

*2) Context-predicting methods*, such as the Collobert and Weston model [25], the hierarchical log-bilinear model (HLBL) [26] and Skip-gram/CBOW models [27]. These kinds of methods are based on the local context window and neural network structures as the underlying predictive model to induce word representations.

III. APPROACHES FOR INTEGRATING WORD REPRESENTATIONS WITH NER

*A. Brown Clustering (BC)*

Brown clustering [17] is an agglomerative hierarchical word clustering technique which group similar words into clusters using the mutual information computed at the bigrams level [1]. The algorithm takes an input sequence $w_1, \ldots, w_n$ of words and returns word clusters as binary tree, where each leaf is an input word. Thus, we can uniquely identify each word by its path from the root. Sample Brown clusters are shown in Table I.

TABLE I
BROWN CLUSTER INDUCED FROM ARABIC WIKIPEDIA DUMPS [28].

| Brown Clusters | Word |
|---|---|
| 010010 | بيلينسكي |
| 010010 | جيرينوفسكي |
| 0000011 | الهيدربروتينية |
| 0000011 | الهيدروفلاحية |
| 0101110 | اليعقوبى |
| 0101110 | العزوزي |
| 1111110001 | بالرباط |
| 1111110001 | بسوريا |

### B. Dense Embeddings (DE)

The intuitive way of integrating word embeddings into NER linear models is the use of dense continuous vector representations of words directly as features. Previous work has demonstrated the effectiveness of this approach to enhance the existing supervised NLP systems [2]. However, it has some disadvantages, such as the problem of linear non-separability, inadequacy when dealing with rare-words and the word ambiguity, and the large amount of computation needed [29]. Figure 1 shows examples of dense embedding features.

| | | | | | | | |
|---|---|---|---|---|---|---|---|
| … | 0.620007 | 0.278912 | 0.812259 | - 0.109514 | 0.093741 | 0.764120 | لاعب |
| … | - 0.306802 | 0.113255 | - 0.425959 | 0.562338 | 0.745610 | 0.847846 | برتغالي |
| … | - 0.160851 | 0.427788 | 0.659257 | - 0.245479 | 0.189593 | - 0.127708 | محترف |

Figure 1. Examples of dense word embeddings.

### C. Binarized Embeddings (BI)

One straightforward way for transforming the low-dimensional continuous-valued word embeddings to high-dimensional discrete embeddings, is binarization.

There are various conversion functions to perform binarization. In this study, we consider three simple ones.

The first function (method A) introduced by Faruqui et al. [30], transforms the real-valued $x_{i,j}$ of the word embedding vector X into binary values by applying:

$$\emptyset(x_{i,j}) = \begin{cases} 1, & if\ x_{i,j} > 0 \\ 0, & otherwise \end{cases}$$

Figure 2 shows examples of the binary embedding features generated using method A.

| | | | | | | | |
|---|---|---|---|---|---|---|---|
| … | 1 | 1 | 0 | 1 | 1 | 1 | لاعب |
| … | 1 | 1 | 0 | 1 | 0 | 1 | برتغالي |
| … | 1 | 0 | 0 | 1 | 1 | 1 | محترف |

Figure 2. Examples of binary word embeddings (method A).

The second function (method B) proposed by Guo et al [5] create binarized embeddings by performing the following conversion:

$$\varphi(x_{i,j}) = \begin{cases} +U, & if\ x_{i,j} \geq MEAN^+ \\ -B, & if\ x_{i,j} \leq MEAN^- \\ 0, & otherwise \end{cases}$$

where $MEAN^+$ ($MEAN^-$) represents the mean of positive (negative) values of vector X.

Figure 3 presents examples of the binary embedding features created using method B.

| | … | +U | +U | 0 | -B | 0 | +U | لاعب |
| --- | --- | --- | --- | --- | --- | --- | --- | --- |
| | … | -B | +U | +U | 0 | 0 | +U | برتغالي |
| | … | +U | 0 | +U | -B | 0 | +U | محترف |

Figure 3. Examples of binary word embeddings (method B).

The third function (method C) that we use is quiet similar to the second one, the only difference is that we calculate the median instead of the mean as follows:

$$\omega(x_{i,j}) = \begin{cases} U_+, & if\ x_{i,j} \geq MEDIAN^+ \\ B_-, & if\ x_{i,j} \leq MEDIAN^- \\ 0, & otherwise \end{cases}$$

where $MEDIAN^+$ ($MEDIAN^-$) represents the median of positive (negative) values of vector X.

Figure 4 presents examples of the binary embedding features generated using method C.

| | … | +U | +U | -B | 0 | 0 | +U | لاعب |
| --- | --- | --- | --- | --- | --- | --- | --- | --- |
| | … | -B | +U | 0 | +U | 0 | +U | برتغالي |
| | … | +U | 0 | 0 | +U | -B | +U | محترف |

Figure 4. Examples of binary word embeddings (method C).

*D. Sparse Embeddings*

The intuition behind this approach is to transform the dense and uninterpretable word embeddings into sparse word vectors. The Introduction of sparsity in word embeddings has been shown to improve usability of word vectors as features [5], dimension interpretability [31], [32] and computational efficiency.

Faruqui et al. [30] introduced two methods to obtain sparse overcomplete word vectors.

The first method is based on sparse coding [33] and $\ell_1$ regularization to create sparse overcomplete word embeddings (SE) as follows:

$$a_{k+1,i,j} = \begin{cases} 0, & if\ |\bar{g}_{k,i,j}| \leq \delta \\ \gamma, & otherwise \end{cases}$$

where, $a_{k+1,i,j}$ is the jth element of overcomplete sparse word vector $a_i$ at the kth update and $\bar{g}_{k,i,j}$ is the corresponding average gradient and $\gamma$ is defined as:

$$\gamma = -sgn(\bar{g}_{k,i,j}) \frac{\eta t}{\sqrt{G_{k,i,j}}} (|\bar{g}_{k,i,j}| - \delta)$$

With $G_{k,i,j} = \sum_{k'=1}^{k} g_{k',i,j}^2$

Figure 5 shows examples of the sparse overcomplete embedding features.

| | | | | | | | |
|---|---|---|---|---|---|---|---|
| … | 0.0525 | 0 | 0 | -0.0315 | 0 | 0 | لاعب |
| … | -0.00938 | 0 | 0 | 0.000644 | 0 | 0 | برتغالي |
| … | 0 | -0.036 | 0 | 0 | 0.0162 | 0 | محترف |

Figure 5. Examples of sparse overcomplete word embeddings.

The second method is based on Non-negative sparse coding and $\ell_1$ regularization to obtain nonnegative sparse word embeddings (NNSE) by zeroing out the negative elements as follows:

$$b_{k+1,i,j} = \begin{cases} 0, & if \ |\bar{g}_{k,i,j}| \leq \delta \\ 0, & if \ \gamma < 0 \\ \gamma, & otherwise \end{cases}$$

where, $b_{k+1,i,j}$ is the jth element of nonnegative sparse word vector $b_i$ at the kth update and $\bar{g}_{k,i,j}$ is the corresponding average gradient and $\gamma$ as defined previously for the first method. Figure 6 shows examples of the nonnegative sparse embedding features.

Both methods have the same hyperparameters: the $\ell_1$-regularization coefficient δ, the $\ell_2$-regularization coefficient τ, and the sparse word embedding length K.

| | | | | | | | |
|---|---|---|---|---|---|---|---|
| … | 0.0527 | 0 | 0 | 0.024 | 0 | 0 | لاعب |
| … | 0.0132 | 0 | 0 | 0.0242 | 0 | 0 | برتغالي |
| … | 0 | 0.00276 | 0 | 0 | 0.0114 | 0 | محترف |

Figure 6. Examples of nonnegative sparse word embeddings.

*E. Clustering Embeddings (CE)*

Yu et al. [29] proposed clustering embeddings to overcome the drawbacks of using the dense embeddings directly with linear models. The k-means clustering technique is used to cluster the word embeddings. The distance metric chosen to measure similarities between clusters and words is the Euclidean distance. Since different granularity can be represented by different numbers of clusters ks [29], we decided to combine the clustering results of different ks as features to fully use the embeddings potential efficiently.

Moreover, based on the cluster features, more discriminative compound features can be built. These compound cluster features are created by combining cluster features internally or with other basic features.

*F. Distributional Prototypes (Proto)*

The distributional Prototypes were introduced by Guo et al [5] for English NER. The basic intuition of these features is that similar words are likely to be labeled with the same entity class. Thus, this approach selects representative words (prototypes) for each class and assigns them as features to the words according to the distributed similarity.

To build distributional prototype features, first, we construct the prototypes list for each target entity class using Normalized Pointwise Mutual Information (NPMI) [34]. The NPMI between entity classes and words from the annotated training corpus is computed as follows:

$$NPMI(class, word) = \frac{PMI(class, word)}{-\ln p(class, word)}$$

$$PMI(class, word) = \ln \frac{PMI(class, word)}{p(class)p(word)}$$

Then we select the top m words for each target class as prototypes.

Table II shows five prototypes extracted from the NER training set of AQMAR corpus [35]

TABLE II
PROTOTYPES EXTRACTED FROM THE AQMAR TRAINING SET USING NPMI.

| Entity Class | Prototypes |
| --- | --- |
| B-LOC | أمريكا, مصر, القدس, دمشق, البرتغال |
| I-LOC | أفريقيا, المتحدة, الجنوبية, المنورة, المقدسة |
| B-PER | محمد, لويس, أحمد, رونالدو, الرازي |
| I-PER | أبي, طولون, بكر, عبد, بن |
| B-ORG | ريال, اتحاد, الفيفا, نادي, منتخب |
| I-ORG | لشبونة, سبورتنغ, مدريد, البرتغالي, يونايتد |
| B-MISC | الميكانيكا, البروتونات, الشابكة, الإلكترونات, اليورانيوم |
| I-MISC | الإخراج, الإدخال, الكلاسيكية, التشغيل, الصليبية |
| O | و, من, في, . . , ، |

Next, given the list of prototypes for each class, the cosine similarity is computed between each word in the corpus and the prototypes in the list using the corresponding word embedding vectors. If the cosine similarity exceeds the predefined threshold (typically 0.5), the prototype will be assigned as a prototype feature of the word. Figure 7 shows examples of assigned prototypes.

| | | | | | | | |
| --- | --- | --- | --- | --- | --- | --- | --- |
| ... | إسبانيا | السنغال | أوقيانوسيا | كوستاريكا | جنوب | فرنسا | **إفريقيا** |
| ... | الصليبيون | إسكندر | الأيوبيين | المستنصر | المنصوري | الرها | **بيبرس** |
| ... | البروتونات | الهيدروجين | الهندسية | الميكانيكا | الإلكترون | النيون | **الموائع** |

Figure 7. Example of prototype features assigned to words.

IV. EXPERIMENTS AND DISCUSSION

*A. NER Model*

In this study, we follow a supervised machine learning approach. Typically, NER is treated as a sequence labeling problem with the aim to find the best label sequence for a given input sequence.

Among the supervised machine learning algorithms, CRF is the most widely used model for sequence labeling in the field of NLP. Hence, we choose it for our NER experiments.

CRF is a discriminative undirected graphical model [36] that integrates the advantages of classification and graphical modeling.

Here, we use the CRFsuite[2] implementation of CRF. It's fast and we can easily change the feature generation code to add extra features.

*B. Baseline Features*

The baseline features were defined over a context window of $\pm 1$ token. The set of features for each token was:
- The word itself.
- Part-of-speech tag.
- Token length.
- Prefixes and Suffixes: the first and last 1, 2, 3, 4 characters in a word.
- Character n-grams: head and trailing character unigrams, bigrams and trigrams.

*C. AQMAR Corpus*

The Arabic Wikipedia Named Entity Corpus (AQMAR) is a hand-annotated corpus of 28 Arabic Wikipedia articles for Arabic named entities [35]. The AQMAR is tagged with four entities: PERSON, LOCATION, ORGANIZATION, MISCELLANEOUS (MIS) and nine classes: O, B-PER, I-PER, B-ORG, I-ORG, B-LOC, I-LOC, B-MISC and I-MISC. In this corpus there are 74K tokens and 2687 sentences. Additional information about AQMAR is shown in Table III.

TABLE III
DEVELOPMENT AND TEST CORPORA STATISTICS FOR AQMAR DATASET.

|  | documents | words | sentences | entities | MIS rate |
|---|---|---|---|---|---|
| Test | 20 | 52,650 | 1,976 | 3,781 | 37% |
| Development | 8 | 21,203 | 711 | 2,073 | 53% |

In this study, we used the test part as the training set. We split the development part in half; one was used as development corpus and the other as testing corpus.

*D. Experimental Setting*

We take the Arabic Wikipedia dumps offered by Al-Rfou', Perozzi and Skiena[3] [28] as our unlabeled data to learn the word embeddings. No pre-processing was done and the text was already tokenized. This corpus contains about 52 million tokens and 1.83 million word types. We set the frequency threshold to 80 and use a dictionary with 48088 most common words in the corpus. Three neural network embedding algorithms were used to learn word embeddings: SKIP-gram, CBOW and GloVe. Table IV presents the training parameters used for each algorithm.

TABLE IV
PARAMETERS USED TO LEARN THE ARABIC WORD EMBEDDINGS

|  | SKIP-gram | CBOW | GloVe |
|---|---|---|---|
| Vector size | 50 | 50 | 50 |
| Window | 5 | 5 | 15 |
| Sample | 1e-3 | 1e-3 | N/A |
| Hierarchical Softmax | Yes | Yes | N/A |
| Negative sampling | 0 | 0 | N/A |
| Frequency threshold | 80 | 80 | 80 |
| Max iterations | N/A | N/A | 15 |
| X_Max | N/A | N/A | 10 |

---

[2] http://www.chokkan.org/software/crfsuite/
[3] https://sites.google.com/site/rmyeid/projects/polyglot#TOC-Download-Wikipedia-Text-Dumps

For the sparse features, we do a grid search on $\delta \in \{0.1, 0.5, 1.0\}$, $\tau \in \{10^{-4}, 10^{-5}, 10^{-6}\}$ and $K \in \{10L, 15L, 20L\}$, where L is the size of the initial vector, to select the three hyperparameters $\delta$, $\tau$ and K empirically. The chosen hyperparameters are summarized in Table V.

For the cluster features, we tune n the number of clusters from 100 to 1000 on the development set, and choose the combination of n =100, 200, 300, 400, 500, 1000, which gets the best results. Concerning the distributional prototype features, we use a fixed number of prototypes m for each target class. m is fine-tuned on the development corpus and set to 60.

Also, we induce 500 brown clusters of words using the same training data as for word embeddings learning.

TABLE V
HYPERPARAMETERS USED FOR LEARNING SPARSE OVERCOMPLETE VECTORS

| X | L | $\delta$ | $\tau$ | K |
|---|---|---|---|---|
| SKIP-gram | 50 | 0.5 | $10^{-5}$ | 500 |
| CBOW | 50 | 0.5 | $10^{-5}$ | 500 |
| GloVe | 50 | 0.5 | $10^{-5}$ | 500 |

*E. Results & Discussion*

Table VI shows the performances of the CRF based NER system on the AQMAR dataset when using different word representation features. The baseline achieves 55.45% of F-score.

Across the three neural network embedding algorithms used, almost all the embedding features improve the baseline. The only exceptions to this rule are sparse embeddings and binary embeddings (method A). The first one does worse than the baseline for SKIP-gram and CBOW. For the second one, It's for SKIP-gram and GloVe.

In comparison with the other embedding features, the Clustered embedding achieves the best performance among the four embedding approaches (4.51% higher than the baseline for SKIP-gram). The second best embedding approach is Distributional Prototypes with an F-score improvement of 1.74% above the baseline for SKIP-gram. The combination of these two features further improves the F-score by 5.75%.

On the other hand, we also compare the embedding features with Brown clustering features. As shown in Table VI, Brown clusters outperform all the embedding features with an F-score of 62.45% (7 points higher than the baseline). Finally, by combining the Brown clustering features with the best embedding features (CE and Proto), the performance can be enhanced further (67.22% for SKIP-gram).

Although the sparse word embeddings presents consistent benefits across many NLP benchmark tasks for English [30], empirical results shows, to our surprise, values close to the baseline and even lower that the ones produced by dense embedding features. This suggests that sparse embedding features are not beneficial enough for the NER task, especially for a morphologically rich language as Arabic.

Generally, our work demonstrates that the integration of the unsupervised word representation features have a very positive impact on NER performance for Arabic. Also, it is worth mentioning that the combination of Brown clusters and various word embeddings approaches further enhance the performances of the Arabic NER system.

TABLE VI
AQMAR NER RESULTS

| | SKIP-gram | CBOW | GloVe |
|---|---|---|---|
| | F1 | F1 | F1 |
| Baseline | 55,45 | 55,45 | 55,45 |
| + DE | 56,18 | 56,19 | 56,41 |
| + SE | 55,26 | 55,36 | 55,51 |
| + NNSE | 55,51 | 55,58 | 55,79 |
| + BI (method A) | 55,01 | 56,87 | 55,26 |
| + BI (method B) | 56,95 | 56,55 | 56,3 |
| + BI (method C) | 56,75 | 55,54 | 56,48 |
| + CE | **59,96** | **60,28** | **60,98** |
| + Proto | 57.19 | 58.9 | 56.86 |
| + CE + Proto | **61,2** | **61,05** | **60,2** |
| + BC | 62,45 | 62,45 | 62,45 |
| + BC + Proto | 63,42 | **66,57** | 63,55 |
| + BC + CE | 65,32 | 64,39 | **66,23** |
| + BC + CE + Proto | **67,22** | **66,57** | 65,36 |

## V. CONCLUSION

This paper investigates the impact of word representations on the Arabic NER system. We present six approaches used for integrating word representations with NER and we provide comparison for three popular neural word embedding algorithms. The Evaluation using AQMAR dataset shows that word representations features boost significantly supervised NER system in Arabic. The performance is further improved when different approaches are combined together.

For future work, we would like to test these approaches with numerous different domains and also investigate the impact of Cross-lingual Word Representations on NER performance for Arabic.